\documentclass{article}

\PassOptionsToPackage{numbers, compress}{natbib}

\usepackage[textsize=footnotesize]{todonotes}

\usepackage[final]{neurips_2024_ml4ps}

\usepackage[utf8]{inputenc} 
\usepackage[T1]{fontenc}    
\usepackage{hyperref}       
\usepackage{url}            
\usepackage{booktabs}       
\usepackage{amsfonts}       
\usepackage{amsmath}
\usepackage{nicefrac}       
\usepackage{microtype}      
\usepackage{xcolor}         
\usepackage{caption}
\definecolor{ok}{rgb}{0,0.8,0}
\definecolor{firstdraft}{rgb}{0,0.6,0}
\definecolor{lena}{rgb}{0.9, 0.5, 0.7}
\definecolor{zhuo}{rgb}{0,0.5,0.5}
\definecolor{towrite}{rgb}{1,0,0}
\definecolor{askira}{rgb}{1,0.33,0.64} 
\definecolor{finalmaybe}{rgb}{0.6,0,0.6}
\definecolor{finalcontentwordingsucks}{rgb}{0.9,0,0.9}

\title{Galaxy Morphology Classification with Counterfactual Explanation}

\author{%
  Zhuo Cao$^1$ \quad Lena Krieger$^1$ \quad Hanno Scharr$^1$ \quad Ira Assent$^{1,2}$ \\
  $^1$IAS-8, Forschungszentrum Jülich, Germany \quad $^2$Aarhus University, Denmark\\
\texttt{\{z.cao, l.krieger, h.scharr, i.assent\}@fz-juelich.de} \\
}

\begin{document}

\maketitle

\begin{abstract}
Galaxy morphologies play an essential role in the study of the evolution of galaxies. 
The determination of morphologies is laborious for a large amount of data giving rise to machine learning-based approaches. 
Unfortunately, most of these approaches offer no insight into how the model works and make the results difficult to understand and explain. 
We here propose to extend a classical encoder-decoder architecture with invertible flow, allowing us to not only obtain a good predictive performance but also provide additional information about the decision process with counterfactual explanations. 
\end{abstract}

\section{Introduction}
Galaxies are the primary building blocks of the universe, composed of stars, stellar remnants, interstellar gas, dust, and dark matter.
A key objective in galaxy research is to elucidate how galaxies have evolved from their early stages to the diverse and large forms observed today \cite{2014ApJ...787..130W}. 
Specifically, analyzing the morphology and structure of galaxies is essential for understanding their evolution, as these aspects are intricately linked to their evolutionary history and are crucial for exploring the physical parameters of galaxies. 
Morphological features are essential for interpreting its evolutionary history and determining a galaxy’s current dynamic state, such as the distribution and movement of stars, gas, and dark matter.

Significant efforts have been dedicated to designing galaxy morphology classification schemes and data collection methods. 
For example, Galaxy Zoo \cite{2008MNRAS.389.1179L} and its successor Galaxy Zoo 2 \cite{galaxyzoo2}, classify galaxies from the Sloan Digital Sky Survey (SDSS) \cite{York_2000} into basic types.
Recently, the classification of galaxy morphologies can be predicted with CNN-based models \cite{2021MNRAS.506..659C, 2024A&A...683A..42C,barchi2020machine, pandya20232}. These automated approaches surpass previous methods and have been applied across multiple surveys \cite{2018MNRAS.476.3661D, 2018ApJ...858..114H, 2020MNRAS.493.4209C}.
The drawback of these methods is their black-box characteristics, limiting the application of these methods because of the lack of interpretability and explainability. 
In this work, we target this issue with validating and insightful counterfactual explanations, demonstrating the importance of certain features for the decision-making process.
\section{Data and Methodology} \label{Sec: data_method}
\paragraph{Visual counterfactual explanations}
Visual counterfactual explanations (CEs) \cite{mothilal2020explaining, wachter2017counterfactual} seek to make only semantically meaningful modifications to an input image in order to obtain a similar image with a target label prediction outcome. 

For a given image $\mathbf{x}$, the objective is to find a counterfactual proposal $\mathbf{x}^{cf}$ that has low counterfactual (CF) loss:
\begin{equation} \label{eq: cf_loss}
    \mathcal{L}_{cf}(\mathbf{x}^{cf}) = f(\mathbf{x}^{cf}, y^{cf}) + s(\mathbf{x}, \mathbf{x}^{cf})
\end{equation}
where a function $s(\mathbf{x}, \mathbf{x}^{cf})$ quantifies the perceptual distance between $\mathbf{x}$ and $\mathbf{x}^{cf}$ and function $f$ yields a lower loss when the classifier predicts a label for the counterfactual that is closer to the target label $y^{cf}$. In other words, a counterfactual explanation reveals what should have been different in $\mathbf{x}$ to observe a diverse outcome with label $y^{cf}$ instead of $y$. Thus, the approach offers deeper insights into the features that significantly contribute to the model's decision-making. CEs accentuate class-relevant features, illustrating how alterations to these features shift the prediction from one class to another. 

While counterfactual explanations provide insights into the differences between predicted classes, generating them presents challenges. First, counterfactuals must align with the data distribution, meaning they need to look realistic within the context of the original dataset. For example, generating a counterfactual image (e.g., turning a cat into a dog) requires a robust generative model capable of maintaining natural, coherent results. Second, irrelevant features must remain unchanged during the generation process. Altering irrelevant or unrelated features can lead to explanations that are misleading or uninformative. For example, if a counterfactual explanation changes the background of an image when the focus should be on the object itself, the explanation may fail to provide useful insights about the model’s decision-making process. Finally, extracting meaningful representations from high-dimensional data, such as images, poses a significant challenge. In such cases, identifying the most relevant features to adjust while preserving the overall structure is difficult due to the complexity of the feature space. In this work, we address these challenges.

\paragraph{Model architecture}
\begin{figure*}[htb]
    \centering
    \includegraphics[width=0.99\textwidth]{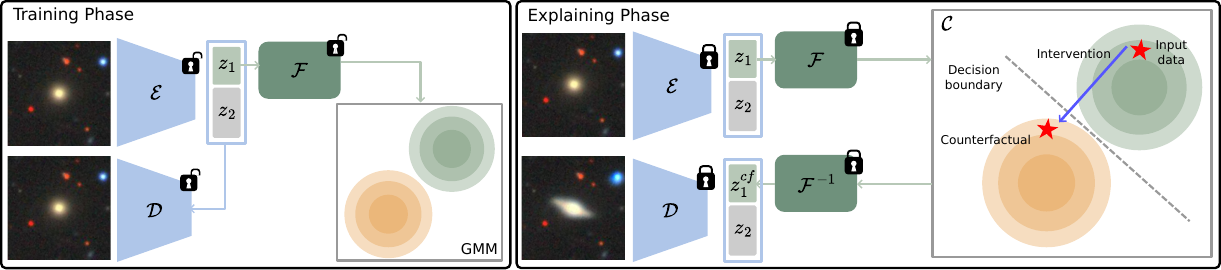}
    \caption{Architecture of our model. Left panel: training phase. Right panel: explanation phase.}
\label{Fig:method}
\end{figure*}

To obtain robust counterfactuals, we construct a model that exploits invertible flows such that counterfactuals obtained in latent space can be translated back to image space.
The proposed model includes three components: an encoder $E$, decoder $D$, and invertible flow $F$ \cite{dinh2016density}.
Briefly, $F$ enables bijective mapping through a specific architecture (see below). As shown in Figure~\ref{Fig:method} (left panel), the encoder maps the image to latent space $\mathcal{Z}$, and the decoder maps latent vectors back to the image space. The invertible flow converts the latent distribution to a Gaussian Mixture Model (GMM) \cite{Reynolds2018GaussianMM}, clustering vectors with the same label in a hidden space $\mathcal{H}$. The input image is classified by the closest cluster mean in $\mathcal{H}$. The entire model can be trained in an end-to-end way.

Compared to commonly used traditional classification methods, the encoder $E$ acts as feature extraction, with an invertible flow replacing MLP. This design reduces the dimensionality of the image data and simplifies the decision boundary to a straight line between two Gaussian means. It also ensures a bijective mapping between latent space $\mathcal{Z}$ and hidden space $\mathcal{H}$, enabling counterfactual explanations. As shown in Figure~\ref{Fig:method} (right panel), the input image maps to a latent vector $\mathbf{z}$, then to a hidden vector $\mathbf{h}$ for classification. A counterfactual latent vector $\mathbf{z}^{cf}$ is created by pushing
$\mathbf{h}$ across the decision boundary and mapping it back to latent space. The decoder $D$ then converts $\mathbf{z}^{cf}$ to image space. The latent space is regularized with Maximum Mean Discrepancy (MMD)~\cite{gretton2006kernel} (see also Equation~\ref{eq:loss}) to keep the encoding function $E$ Lipschitz continuous, ensuring interpolability and meaningful counterfactuals \cite{zhao2017infovae}.

This intervention enables us to achieve an arbitrarily low $f$ loss in Equation~\ref{eq: cf_loss}. However, minimizing the distance between the original and counterfactual images $s(\mathbf{x}, \mathbf{x}^{cf})$ is still necessary, which translates to minimizing the distance between their latent vectors due to the Lipschitz continuity of the encoding function. We achieve this by splitting the latent vector into class-dependent $\mathbf{z}_1$ and class-independent $\mathbf{z}_2$ components, where only $\mathbf{z}_1$ is used by the invertible flow $F$ for classification. Additionally, we train the invertible flow with an information bottleneck objective \cite{ardizzone2020training} (see also Equation~\ref{eq:loss} and \ref{eq:loss_ib}) to reduce mutual information between latent and hidden spaces, ensuring that $F$ focuses on essential classification information with minimal alteration to $\mathbf{z}_1$ similar to pixel-level counterfactual generation by \cite{hvilshoj2021ecinn}. Note that the decoder may ignore $\mathbf{z_1}$ completely if $\mathbf{z_2}$ contains duplicated information to $\mathbf{z_1}$. Applying MMD constraints in the latent space removes the redundancy in the latent vector \cite{higgins2017betavae}, making the generated image respond to the modification in $\mathbf{z_1}$.
The details of the model and training procedure can be found in Appendix~\ref{app:exp_set}.
\paragraph{Invertible Flow}
An Invertible flow, or Invertible Neural Network (INN) \cite{dinh2016density,dinh2014nice,kingma2018glow}, is a type of neural network architecture designed so that its forward and backward operations are both computationally feasible and reversible. This means that given the output, the original input can be accurately reconstructed. INNs achieve this by using specific structures that ensure bijective (one-to-one) mappings between the input and output spaces, called coupling layers. In a coupling layer, the input data is split into two parts. One part of the data remains unchanged, while the other part is transformed using a function conditioned on the first part. This approach ensures that the transformation is invertible and the Jacobian determinant of the transformation is easy to compute. 

\paragraph{Loss Functions} \label{sec:loss_func}
As previously mentioned, the objective comprises two key aspects: firstly, the model is required to generate in-distribution images, and secondly, decision-irrelevant features should remain unchanged. To accomplish this, we design the loss functions as follows:
\begin{equation}
    \mathcal{L} = \mathcal{L}_{\mathrm{R}}(\mathbf{\tilde{x}}, \mathbf{x}) + \mathcal{L}_{\mathrm{MMD}}(\mathbf{z},\mathbf{n}) + \mathcal{L}_{\mathrm{IB}}(\mathbf{z_1}, y)\label{eq:loss}
\end{equation}
Here, $\mathcal{L}_\mathrm{R}(\mathbf{\tilde{x}}, \mathbf{x}) = \Phi(\mathbf{\tilde{x}}) - \Phi(\mathbf{x})$ denotes the reconstruction loss between the generated image ($\mathbf{\tilde{x}}$) and the input image ($\mathbf{x}$), where $\Phi$ represents a VGG16 model pre-trained on the ImageNet dataset \cite{5206848}. This model captures meaningful representations, as discussed in \cite{johnson2016perceptual}.

The second term, $\mathcal{L}_\mathrm{MMD}$, describes the Maximum Mean Discrepancy loss, which encourages the latent vector ($\mathbf{z}$) to approximate a Gaussian distribution. This ensures the interpolability of the latent space so that the generated image with the modified latent vector is still in-distribution. The loss is empirically estimated \cite{gretton2006kernel} by 
\begin{align}
    \mathcal{L}_{\mathrm{MMD}}(\mathbf{z},\mathbf{n})
    & = \frac{1}{m(m-1)} \sum_{i=1}^{m} \sum_{j \neq i} k(\mathbf{z}^i, \mathbf{z}^j) 
    - \frac{2}{m^2} \sum_{i=1}^{m} \sum_{j=1}^{m} k(\mathbf{z}^i, \mathbf{n}^j) \\ \label{eq:MMD}
    & + \frac{1}{m(m-1)} \sum_{i=1}^{m} \sum_{j \neq i} k(\mathbf{n}^i, \mathbf{n}^j) \nonumber
\end{align}
where $m$ is the batch size while training. The term $\mathbf{z}^i= E(\mathbf{x}^i)$ denotes the latent vector of $i$-th input data $\mathbf{x}^i $, $\mathbf{n}^i$ represents the $i$-th sample from the target Gaussian distribution, and $k$ is the kernel, more precisely a Radial Basis Function (RBF) in this work.
Together with the reconstruction loss $\mathcal{L}_\mathrm{R}$, these components constitute the MMD-VAE \cite{zhao2017infovae}, a generative model known for its strong reconstruction quality. 

The last term in the loss function $\mathcal{L}$ (Equation~\ref{eq:loss}) represents the information bottleneck loss \cite{ardizzone2020training, tishby2000information}. On high-level, it is expressed as:
\begin{equation}
    \mathcal{L}_\mathrm{IB}
     = I(\mathcal{Z}, \mathcal{H}) - \beta I(\mathcal{H}, \mathcal{Y})\label{eq:loss_ib}
\end{equation}
The first term in Equation~\ref{eq:loss_ib} on the right-hand side minimizes the mutual information $I$ between the latent space $\mathcal{Z}$ and the hidden space $\mathcal{H}$, while the second term maximizes the mutual information between the hidden space and the class label $\mathcal{Y}$. The combination of both components ensures that only essential information is used for classification. The parameter $\beta$ controls the trade-off between preserving relevant information and discarding irrelevant details. Higher $\beta$ values emphasize task performance, leading to better classification accuracy but potentially less robust uncertainty quantification.
Lower $\beta$ values prioritize compression, resulting in improved uncertainty calibration and out-of-distribution detection, at the cost of some classification accuracy. In this work, an intermediate value of 3 is used. For the detailed implementation of this loss function, readers are referred to the original paper \cite{ardizzone2020training}. Note that this loss connects the input data $\mathbf{x}$ to its corresponding class label $y$ and is applied only to the class-dependent component $\mathbf{z}_1$.

\begin{figure}[h]
  \centering
  \includegraphics[width =0.8\textwidth ]{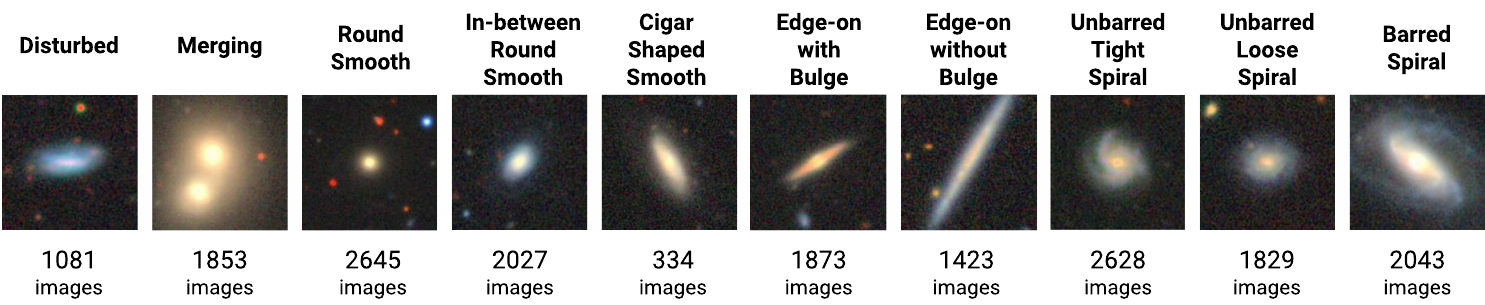}
  \caption{Sample images for each class of Galaxy 10 DECaLS and number of instances.}\label{Fig:data_demo}
\end{figure}

\paragraph{Galaxy10 DECaLS}
We study Galaxy 10 DECaLS\footnote{The data is publicly available \href{https://astronn.readthedocs.io/en/latest/galaxy10.html}{online}} with 17,736 DESI Legacy Imaging Surveys (DECaLS)~\cite{dey2019overview} images (g, r and z band) and labels from Galaxy Zoo Release 2~\cite{galaxyzoo2} describing galaxy morphology in ten distinct classes. Figure~\ref{Fig:data_demo} shows examples of each class.
We normalize the images from [0, 255] to [0, 1], rotate the image randomly, and resize it to 256 x 256 after cropping the central region.


\section{Results}
\paragraph{Metrics}
We evaluate the model with regard to accuracy and similarity. Our trained model reports an accuracy of $\sim80$\%. The accuracy for each class varies between 70\% to 90\% except for the disturbed galaxy, which has an accuracy of about 41\% reflecting its complexity. The similarity between the counterfactual and original images is high, with Mean Squared Distance of $0.006$ and Structural Similarity Index Measure (SSIM) \cite{1284395} of $0.96$. SSIM is designed to better align with human visual perception compared to traditional
metrics. Details in Appendix~\ref{app: metric_analysis}.
\begin{figure}[htb]
    \centering
    \includegraphics[width=0.95\textwidth]{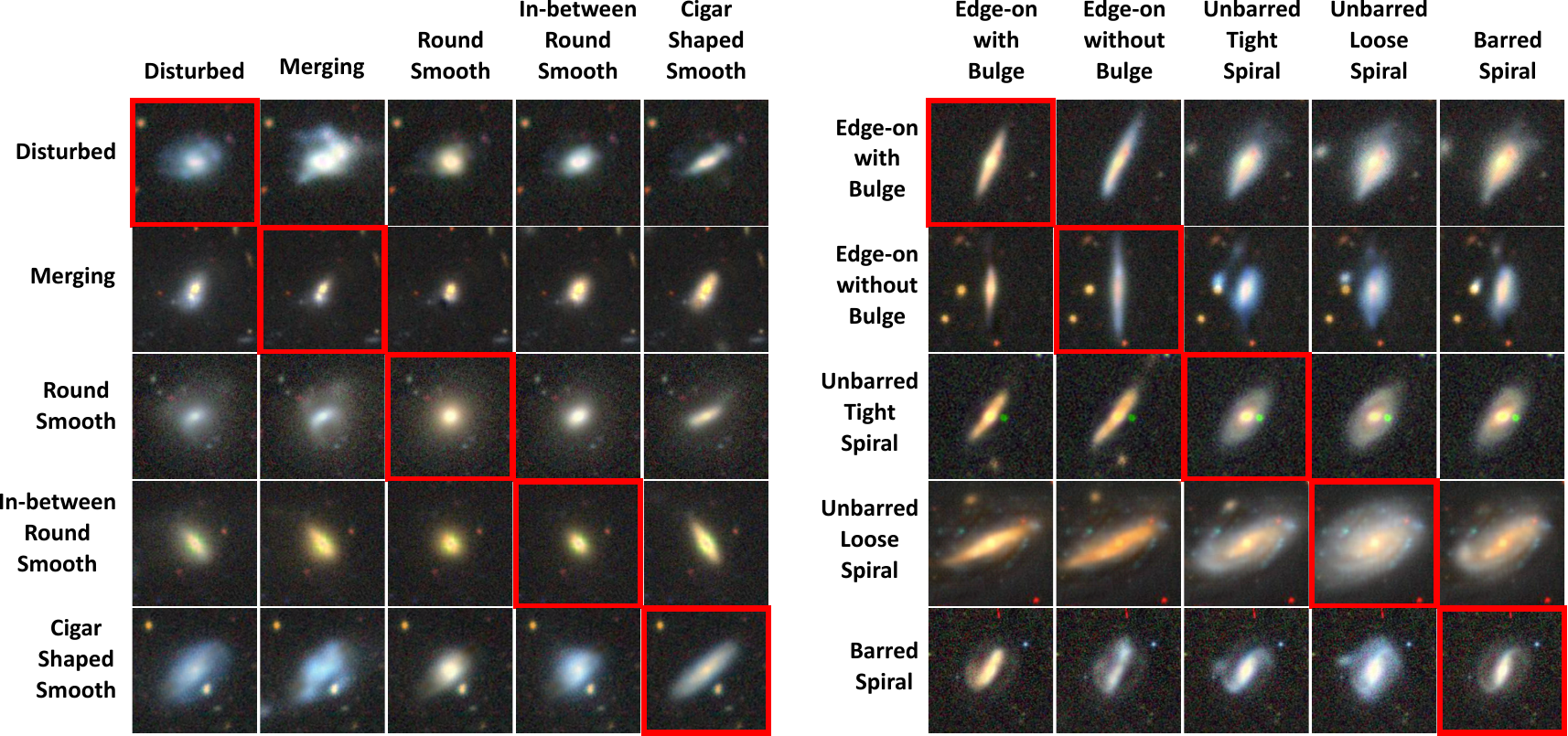}
    \caption{Original images aligned across diagonal (red boxes) with counterfactuals displayed in the same row classified according to column labels. Due to limited space, a complete image grid is shown in Appendix \ref{app: complete_visualization}.}
\label{Fig:cf_example}
\end{figure}
\paragraph{Counterfactual Explanation}
\label{subsec:vis_results}
Results are visualized in Figure~\ref{Fig:cf_example} by overlaying the difference between counterfactual $\mathbf{x}^{cf}$ and reconstructed image $\mathbf{\tilde{x}}$ on the input image $\mathbf{x}$, i.e. $\mathbf{x} + (\mathbf{x}^{cf} - \mathbf{\tilde{x}})$. Notably, decision-relevant areas can be emphasized as in other Explainable AI methods, by calculating the difference between the counterfactual and the reconstruction, $\mathbf{x}^{cf} - \mathbf{\tilde{x}}$ (Example in Appendix~\ref{app: complete_visualization}).

When comparing round and cigar-shaped smooth galaxies, we notice the change in shape from round to more elongated as expected.
Similarly, edge-on galaxies with and without bulge are clearly distinguishable by the central galaxy bulge.
Barred and unbarred spiral galaxies can be distinguished by their central structures, with barred spirals featuring elongated central objects and unbarred spirals having rounder central regions.
Please note that there is no change in background during reconstruction. 
Thus it can be concluded that the encoder correctly distinguishes between class-dependent and class-independent latent features.

\begin{figure}[htb]
    \centering
    \includegraphics[width=0.9\textwidth]{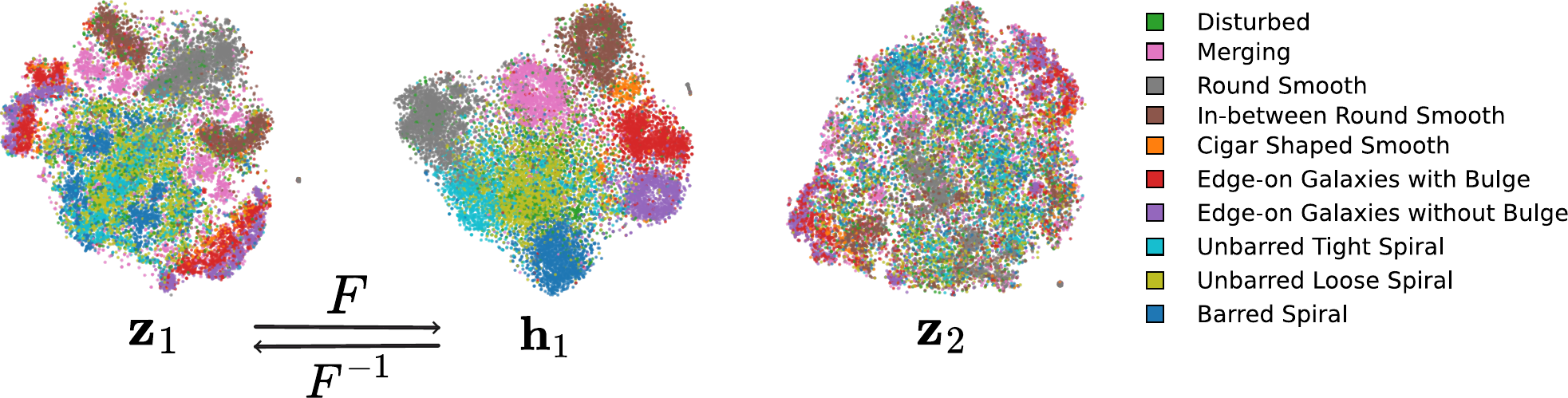}
    \caption{t-SNE plots for class-dependent ($\mathbf{z}_1$), hidden ($\mathbf{h}_1$) and background ($\mathbf{z}_2$) features.}
\label{Fig:tsne_plot}
\end{figure}

We observe that the spiral morphologies, i.e., barred tight/ loose and unbarred spiral, and their visual counterfactual explanations, displayed in the right panel, look very similar. 
As the image is compressed in feature space before classification, the fine structures are gradually removed. 
Therefore, the invertible flow and the decoder have no information about fine structures like spirals. 
The remaining information is likely shared in their latent features. 
As a result, the groups are close to each other in the latent space and the distance to the nearest sample of the neighboring class is very small (see Figure~\ref{Fig:tsne_plot}), corresponding to a minimal difference in the image. 

\paragraph{Latent Space Analysis} The latent space learned by our model is further analyzed with t-SNE plots \cite{van2008visualizing}, see Figure~\ref{Fig:tsne_plot}. 
The left and right t-SNE plots illustrate the latent space features $\mathbf{z}_1$ and $\mathbf{z}_2$. While $\mathbf{z}_1$ shows at least a few connected groups, $\mathbf{z}_2$ hardly shows any groupings corresponding to the classes. 
This is a desired behavior since the information in $\mathbf{z}_2$ should be independent of the classifications.
Hidden space features $\mathbf{h}_1$, i.e., $\mathbf{z}_1$ transformed with $F$ contain mostly clearly separable clusters, indicating a separation of the latent space features regarding their classes. 
The clusters that cannot be separated are Unbarred Loose Spiral (green) and Unbarred Tight Spiral (cyan). 
As already observed above, the model is challenged regarding very fine details in the images, e.g., spirals. The remaining similarities of the classes reflect on the positioning of the clusters in latent space. 

The t-SNE plot does not reveal any information about samples belonging to the Disturbed (dark green) class. These galaxies tend to be diffuse and resemble different classes, rendering them hard to group. This is reflected in the accuracy of 41\% for Disturbed galaxies. 
Even though there are few orange points in $\mathbf{z}_1$'s t-SNE plot due to the few included Cigar Shaped Smooth samples in the dataset, there is a well-visible cluster formed in the t-SNE plot according to $\mathbf{h}_1$. 
This demonstrates the model's ability to handle imbalanced classes.

\section{Discussion}
Previous works employed Explainable AI techniques to astrophysical use cases, such as identifying informative latent space representations of galaxy spectra with SHAP values \cite{iwasaki2023extracting} or detecting ultra-compact dwarfs and globular clusters using Localized Generalized Matrix Learning Vector Quantization (LGMLVQ) to provide feature importance for each class, class-wise representative samples and the possibility for non-linear visualization of the data \cite{mohammadi2022detection}. 
Bhambra et al. \cite{bhambra2022explaining} explain galaxy morphology classification with saliency maps. In contrast to our approach, they apply SmoothGrad to illustrate which pixels contribute to classification. 
Their findings show that the trained ensemble, consisting of the three architectures VGG16, ResNet50v2, and Xception, sometimes disagrees with the target labels in Galaxy Zoo assigned by citizen science. 
They show examples indicating that the ensemble might be more correct than the ground truth.

Our approach is sensitive to mislabeled samples, as the latent space feature vectors are changed and thus the bias is shifted towards the counterfactuals. 
This effect can be investigated by analyzing the distributions determined by invertible flow $F$. 
Closer examination of $\mathbf{h}_1$, see Figure~\ref{Fig:tsne_plot}, can identify problematic classes for further reviewing. This is a desirable property as it identifies possible issues of the pipeline, i.e., data or model, that might not be discovered otherwise.

\section{Conclusion}
We present a new approach to produce realistic counterfactual explanations for galaxy morphologies by adjusting the class-dependent latent space features. 
In future work, we aim to address the current limitation related to capturing fine details within the images. 
Additionally, we are interested in exploring the interpretability that the distributions inside $F$ hold, as this could offer valuable insights into the relationships between the classes and reveal any limitations of the classifier or the data when the classes cannot be clearly distinguished from each other.

\begin{ack}
The authors gratefully acknowledge computing time on the supercomputer JURECA \cite{JURECA} at Forschungszentrum Jülich under grant delia-mp.

\end{ack}

\medskip

{
\small
\bibliographystyle{ieeetr}
\bibliography{literature}
}

\newpage
\appendix
\section{Appendix / Supplemental Material}
\subsection{Experiment setting}\label{app:exp_set}
The Encoder-Decoder architecture can be found in Table~\ref{table:model_architecture}. The output feature has a size of 32, which is split into $\mathbf{z}_1$ and $\mathbf{z}_2$ of lengths 24 and 8, respectively.

The invertible flow is implemented with \texttt{FrEIA} \cite{freia}. The architecture has 32 All-in-One blocks, each containing the following sequences: Linear(24, 128) $\rightarrow$ BN $\rightarrow$ LeakyRelu $\rightarrow$ Linear(128, 128) $\rightarrow$ BN $\rightarrow$ LeakyRelu  $\rightarrow$ Linear(128, 24).

The model is trained using a learning rate of $0.0001$ with an exponential learning rate scheduler, whose factor of learning rate decay is $\gamma=0.99$. The entire dataset is split into training, validation, and test sets with ratios of 0.7, 0.2, and 0.1, respectively. The training batch size is 128 and the number of epochs is 100. The model is trained on a single A100 80GB GPU. 
\begin{table}[htbp]
\centering
\resizebox{\textwidth}{!}{%
\begin{tabular}{lcc|lcc}
\toprule
\textbf{Layer (Encoder)} & \textbf{Filter/Units} & \textbf{Output Shape} & \textbf{Layer (Decoder)} & \textbf{Filter/Units} & \textbf{Output Shape} \\ \midrule
\textsc{Input}          & -               & 256 x 256 x 3       & \textsc{Input}               & -                     & 32         \\
\textsc{DoubleConv1}    & 8               & 256 x 256 x 8       & \textsc{FC1}                 & 2048                  & 2048         \\
\textsc{MaxPool1}       & -               & 128 x 128 x 8       & \textsc{Reshape}             & -                     & 4 x 4 x 256         \\
\textsc{6 x ResBlock1}  & 8               & 128 x 128 x 8       & \textsc{Upsample1}           & 256                   & 8 x 8 x 256         \\

\textsc{DoubleConv2}    & 16              & 128 x 128 x 16      & \textsc{DoubleConv1}         & 128                   & 8 x 8 x 128         \\
\textsc{MaxPool2}       & -               & 64 x 64 x 16        & \textsc{6 x ResBlock1}       & 128                   & 8 x 8 x 128         \\

\textsc{6 x ResBlock2}  & 16              & 64 x 64 x 16        & \textsc{Upsample2}           & 128                   & 16 x 16 x 128         \\
\textsc{DoubleConv3}    & 32              & 64 x 64 x 32        & \textsc{DoubleConv2}         & 64                    & 16 x 16 x 64         \\
\textsc{MaxPool3}       & -               & 32 x 32 x 32        & \textsc{6 x ResBlock2}       & 64                    & 16 x 16 x 64         \\

\textsc{6 x ResBlock3}  & 32              & 32 x 32 x 32        & \textsc{Upsample3}           & 64                    & 32 x 32 x 64         \\
\textsc{DoubleConv4}    & 64              & 32 x 32 x 64        & \textsc{DoubleConv3}         & 32                    & 32 x 32 x 32         \\
\textsc{MaxPool4}       & -               & 16 x 16 x 64        & \textsc{6 x ResBlock3}       & 32                    & 32 x 32 x 32         \\

\textsc{6 x ResBlock4}  & 64              & 16 x 16 x 64        & \textsc{Upsample4}           & 32                    & 64 x 64 x 32         \\
\textsc{DoubleConv5}    & 128             & 16 x 16 x 128       & \textsc{DoubleConv4}         & 16                    & 64 x 64 x 16         \\
\textsc{MaxPool5}       & -               & 8 x 8 x 128         & \textsc{6 x ResBlock4}       & 16                    & 64 x 64 x 16         \\

\textsc{6 x ResBlock5}  & 128             & 8 x 8 x 128         & \textsc{Upsample5}           & 16                    & 128 x 128 x 16         \\
\textsc{DoubleConv6}    & 256             & 8 x 8 x 256         & \textsc{DoubleConv5}         & 8                     & 128 x 128 x 8         \\
\textsc{MaxPool6}       & -               & 4 x 4 x 256         & \textsc{6 x ResBlock5}       & 8                     & 128 x 128 x 8         \\

\textsc{6 x ResBlock6}  & 256             & 4 x 4 x 256         & \textsc{Upsample6}           & 8                     & 256 x 256 x 8         \\
\textsc{Flatten}        & -               & 2048                & \textsc{DoubleConv6}         & 8                     & 256 x 256 x 8         \\
\textsc{FC1}            & 32              & 32                  & \textsc{6 x ResBlock6}       & 8                     & 256 x 256 x 8         \\
                        &                 &                     & \textsc{Conv\&Sigmoid}       & 3                     & 256 x 256 x 3         \\ 
\bottomrule
\end{tabular}
}
\caption{Layer-wise architecture of the encoder and decoder. The \textsc{DoubleConv} layer consists of two sequences of Convolution, Batch Normalization, and ReLU activation layers. A \textsc{ResBlock} comprises the following sequence: Conv(1) $\rightarrow$ ReLU $\rightarrow$ Conv(3) $\rightarrow$ ReLU $\rightarrow$ Conv(3) $\rightarrow$ ReLU $\rightarrow$ Conv(1). Upsampling is performed using linear interpolation.}
\label{table:model_architecture}
\end{table}

\newpage
\subsection{Quantitative Analysis} \label{app: metric_analysis}
\begin{figure}[htbp]
    \centering
    \includegraphics[width=0.75\textwidth]{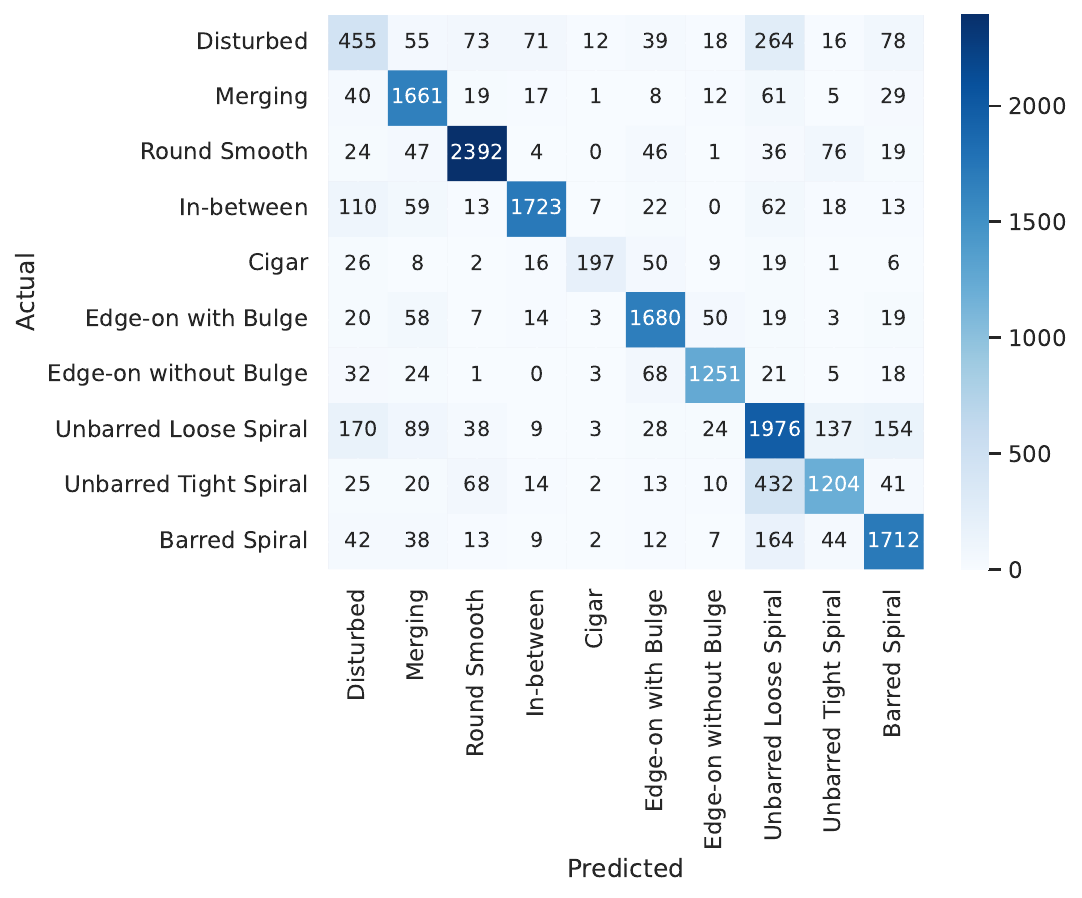}
    \caption{Confusion matrix of the trained model for the entire dataset.}
\label{Fig:conffusion_matrix}
\end{figure}

The classification report and the confusion matrix can be found in Table~\ref{table:classification_report} and Figure~\ref{Fig:conffusion_matrix}.

Based on the results, the model performs best for the round smooth class, achieving an F1-Score of 0.93 (and 0.91 for the entire dataset). Conversely, the disturbed class has the lowest performance, with an F1-Score of 0.41 (0.45 for the entire dataset). This outcome is expected since the morphology of a round smooth galaxy is relatively simple, resembling a compact object, whereas a disturbed galaxy's structure is much more complex. The confusion matrix indicates that disturbed galaxies are frequently misclassified as unbarred loose spiral galaxies. Additionally, unbarred loose and tight spiral galaxies are often difficult to distinguish from one another. 

\begin{table}[htbp]
\centering
\begin{tabular}{lcccc}
\toprule
\textbf{Class} & \textbf{Precision} & \textbf{Recall} & \textbf{F1-Score} & \textbf{Support} \\ \midrule
Disturbed              & 0.40 (0.48) & 0.41 (0.42) & 0.41 (0.45) & 94 (1081) \\
Merging                & 0.79 (0.81) & 0.90 (0.90) & 0.84 (0.85) & 188 (1853) \\
Round Smooth           & 0.94 (0.91) & 0.93 (0.90) & 0.93 (0.91) & 254 (2645) \\
In-between             & 0.91 (0.92) & 0.85 (0.85) & 0.88 (0.88) & 200 (2027) \\
Cigar                  & 0.88 (0.86) & 0.54 (0.59) & 0.67 (0.70) & 28 (334) \\
Edge-on with Bulge     & 0.87 (0.85) & 0.91 (0.90) & 0.89 (0.88) & 191 (1873) \\
Edge-on without Bulge  & 0.94 (0.91) & 0.82 (0.88) & 0.88 (0.89) & 147 (1423) \\
Unbarred Loose Spiral  & 0.64 (0.65) & 0.75 (0.75) & 0.69 (0.70) & 277 (2628) \\
Unbarred Tight Spiral  & 0.79 (0.80) & 0.63 (0.66) & 0.70 (0.72) & 185 (1829) \\
Barred Spiral          & 0.81 (0.82) & 0.82 (0.84) & 0.81 (0.83) & 209 (2043) \\
\midrule
\textbf{Accuracy}      & & & 0.80 (0.80) & \\
\textbf{Macro Avg}     & 0.80 (0.80) & 0.75 (0.77) & 0.77 (0.78) & 1773 (17736) \\
\textbf{Weighted Avg}  & 0.81 (0.81) & 0.80 (0.80) & 0.80 (0.80) & 1773 (17736) \\ \bottomrule
\end{tabular}
\caption{Classification report showing precision, recall, f1-score, and support for different classes. The numbers in and out of the parentheses are for the entire dataset and test set, respectively.}
\label{table:classification_report}
\end{table}

\newpage


\subsection{Complete Counterfactual Visualization} \label{app: complete_visualization}
\begin{figure}[h!]
    \centering
    \includegraphics[width=0.95\textwidth]{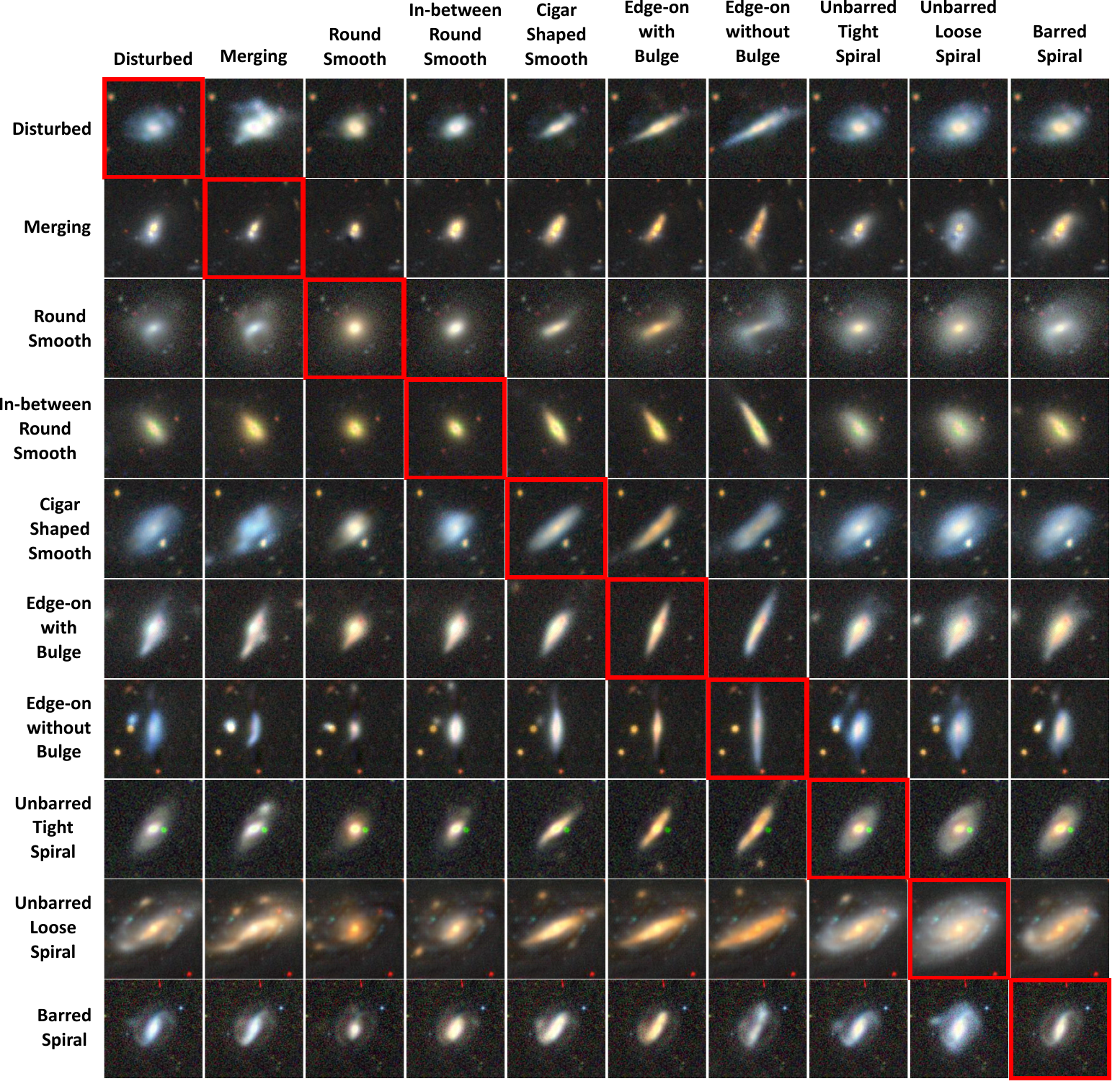}
    \caption{The original images are aligned across the diagonal (red boxes), and the counterfactual images are displayed in the same row their classification according to the labels above.}
\label{Fig:cf_example_complete}
\end{figure}

The complete version of the counterfactual visualization is illustrated in Figure~\ref{Fig:cf_example_complete}. The diagonal entries display the original input images from different galaxy classes, while the off-diagonal entries illustrate how these images are transformed into counterfactuals that resemble other galaxy classes. For instance, in the first row, the original "Disturbed" galaxy is modified to appear as though it belongs to the "Merging," "Round Smooth," "In-between Round Smooth," and other classes. It demonstrates the changes required for each galaxy to be reclassified as a different type, offering insights into the decision boundaries and the alterations necessary to cross them. The diversity and realism of the transformations reflect the model's understanding of the morphological features that differentiate galaxy classes.

\newpage
To emphasize the differences between the original and counterfactual images, we superimpose the difference heatmap $\mathbf{x}^{cf} - \mathbf{\tilde{x}}$  onto the counterfactuals. This approach further highlights the characteristic features of each class. For instance, converting any class to a 'Round Smooth' galaxy (shown in the third column) involves reducing the surrounding details while enhancing the central object, indicating that round smooth galaxies are compact in nature.
\begin{figure}[h!]
    \centering
    \includegraphics[width=0.95\textwidth]{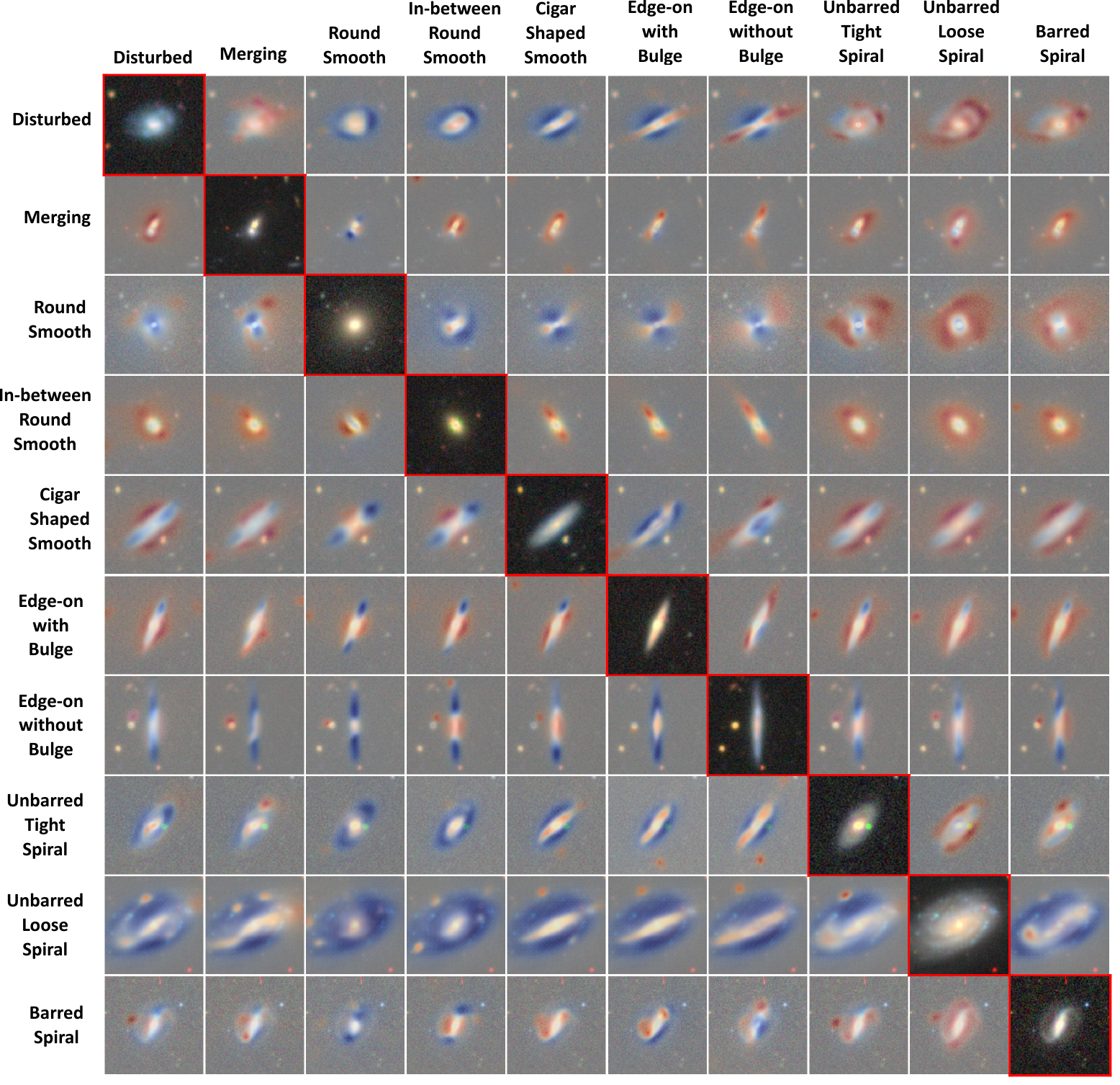}
    \caption{The original images are arranged along the diagonal (highlighted in red boxes), while the counterfactual images are shown in the same row corresponding to their classification as indicated by the labels above. A difference heatmap $\mathbf{x}^{cf} - \mathbf{\tilde{x}}$ is overlayed on the counterfactual images. Reddish colors represent positive values, while bluish colors indicate negative values.}
\label{Fig:cf_example_heatmap}
\end{figure}
\newpage

\end{document}